\documentclass[11pt,a4paper]{article}
\usepackage[hyperref]{acl2021}
\usepackage{times}
\usepackage{latexsym}
\usepackage{spverbatim}

\usepackage{times}
\usepackage{soul}
\usepackage[utf8]{inputenc}
\usepackage{amsfonts}       
\usepackage{nicefrac}       
\usepackage{microtype}      
\usepackage{hhline}
\usepackage{boxedminipage}
\usepackage{hyperref}
\usepackage{amssymb}
\usepackage{wrapfig,lipsum,booktabs}
\usepackage{amsmath}
\usepackage{cleveref}
\usepackage{booktabs}
\usepackage{bm}
\usepackage{algorithm}
\usepackage{algpseudocode}
\usepackage{makecell}
\usepackage{graphicx}
\usepackage{nccmath}
\usepackage{caption}
\usepackage{multirow}
\usepackage{makecell}
\usepackage{subcaption}
\usepackage{tabularx}
\usepackage{colortbl}

\usepackage{microtype}
\PassOptionsToPackage{hyphens}{url}\usepackage{hyperref}

\aclfinalcopy 

\newcommand{\B}[1] {\boldsymbol{#1}}

\def\bb{{\B{b}}}

\def\bu{{\B{u}}}
\def\bv{{\B{v}}}

\def\bx{{\B{x}}}

\title{Fusing Context Into Knowledge Graph for Commonsense Question Answering}

\makeatletter
\newcommand{\thickhline}{%
    \noalign {\ifnum 0=`}\fi \hrule height 1pt
    \futurelet \reserved@a \@xhline
}
\makeatother

\author{Yichong Xu\thanks{$\;\;$Equal contribution}$\;$, Chenguang Zhu$^*$, Ruochen Xu, Yang Liu, Michael Zeng, Xuedong Huang \\
Microsoft Cognitive Services Research Group\\
  \texttt{\{yicxu,chezhu,ruox,yaliu10,nzeng,xdh\}@microsoft.com} \\}
\date{}

\begin{document}
\maketitle
\begin{abstract}
Commonsense question answering (QA) requires a model to grasp commonsense and factual knowledge to answer questions about world events. Many prior methods couple language modeling with knowledge graphs (KG). However, although a KG contains rich structural information, it lacks the context to provide a more precise understanding of the concepts. This creates a gap when fusing knowledge graphs into language modeling, especially when there is insufficient labeled data.
Thus, we propose to employ external entity descriptions to provide contextual information for knowledge understanding. 
We retrieve descriptions of related concepts from Wiktionary and feed them as additional input to pre-trained language models. The resulting model achieves state-of-the-art result in the CommonsenseQA dataset and the best result among non-generative models in OpenBookQA.
Our code is available at \url{https://github.com/microsoft/DEKCOR-CommonsenseQA}.
\end{abstract}

\section{Introduction}
\label{sec:intro}
One critical aspect of human intelligence is the ability to reason over everyday matters based on observation and knowledge. This capability is usually shared by most people as a foundation for communication and interaction with the world.
Therefore, commonsense reasoning has emerged as an important task in natural language understanding, with various datasets and models proposed in this area \citep{ma2019towards,csqa,wang2020connecting,lv2020graph}.

While massive pre-trained models \citep{bert,roberta} are effective in language understanding, they lack modules to explicitly handle knowledge and commonsense. Also, structured data like knowledge graph is much more efficient in representing commonsense compared with unstructured text.
Therefore, there have been multiple methods coupling language models with various forms of knowledge graphs (KG) for commonsense reasoning, including knowledge bases \citep{sap2019atomic,yu2020survey}, relational paths \citep{kagnet}, graph relation network \citep{feng2020scalable} and heterogeneous graph \citep{lv2020graph}. These methods combine the merits of language modeling and structural knowledge information and improve the performance of commonsense reasoning and question answering. 

However, there is still a non-negligible gap between the performance of these models and humans. One reason is that, although a KG can encode topological information between the concepts, it lacks rich context information. For instance, for a graph node for the entity ``Mona Lisa'', the graph depicts its relations to multiple other entities. But given this neighborhood information, it is still hard to infer that it is a painting. On the other hand, we can retrieve the precise definition of ``Mona Lisa'' from external sources, e.g. the definition of Mona Lisa in Wiktionary is ``\textit{A painting by Leonardo da Vinci, widely considered as the most famous painting in history}''. To represent structured data that can be seamlessly integrated into language models, we need to provide a panoramic view of each concept in the knowledge graph, including its neighboring concepts, relations to them, and a definitive description of it.

Thus, we propose the DEKCOR model, i.e. DEscriptive Knowledge for COmmonsense question answeRing, to tackle multiple choice commonsense questions. Given a question and a choice, we first extract the contained concepts. Then, we extract the edge between the question concept and the choice concept in ConceptNet \citep{conceptnet}. If such an edge does not exist, we compute a relevance score for each knowledge triple (subject, relation, object) containing the choice concept, and select the one with the highest score. Next, we retrieve the definition of these concepts from Wiktionary via multiple criteria of text matching. Finally, we feed the question, choice, selected triple and definitions into the language model ALBERT \citep{ALBERT} to produce a score indicating how likely this choice is the correct answer.

We evaluate our model on CommonsenseQA \citep{csqa} and OpenBookQA \citep{mihaylov2018can}. On CommonsenseQA, it outperforms the previous state-of-the-art result by 1.2\% (single model) and 3.8\% (ensemble model) on the test set. On OpenBookQA, our model outperforms all baselines other than two large-scale models based on T5 \citep{t5}. We further conduct ablation studies to demonstrate the effectiveness of fusing context into the knowledge graph.

\section{Related work}
\label{sec:rw}
Several different approaches have been investigated for leveraging external knowledge sources to answer commonsense questions. 
\citet{min2019knowledge} addresses open-domain QA by retrieving from a passage graph, where vertices are passages and edges represent relationships derived from external knowledge bases and co-occurrence. \citet{sap2019atomic} introduces the ATOMIC graph with 877k textual descriptions of inferential knowledge (e.g. if-then relation) to answer causal questions. \citet{kagnet} projects questions and choices to the knowledge-based symbolic space as a schema graph. It then utilizes path-based LSTM to give scores. \citet{feng2020scalable} adopts the multi-hop graph relation network (MHGRN) to perform reasoning unifying path-based methods and graph neural networks. 
\citet{lv2020graph} proposes to extract evidence from both structured knowledge base such as ConceptNet and Wikipedia text and conduct graph-based representation and inference for commonsense reasoning. \citet{wang2020connecting} employs GPT-2 to generate paths between concepts in a knowledge graph, which can dynamically provide multi-hop relations between any pair of concepts.

Several studies have utilized knowledge descriptions for different tasks. \citet{yu2020jaket} uses description text from Wikipedia for knowledge-text co-pretraining. \citet{xie2016representation} encodes the semantics of entity descriptions in knowledge graphs to improve the performance on knowledge graph completion and entity classification. \citet{chen2018co} co-trains the knowledge graph embeddings and entity description representation for cross-lingual entity alignment. Concurrent with our work, \citet{chen-etal-2020-improving-commonsense} also insert knowledge descriptions into commonsense question answering. Compared with our work, the proposed method in \citet{chen-etal-2020-improving-commonsense} is much more complex, e.g. involving training additional rankers on retrieved text, while our result outperforms Chen et al. on CommonsenseQA.

\section{Method}
\label{sec:method}
\subsection{Knowledge Retrieval}
\label{sec:form}
\textbf{Problem formulation.} In this paper, we focus on the following QA task: given a commonsense question $q$, select the correct answer from several choices $c_1, ..., c_n$. In most cases, the question does not contain any mentions of the answer. Therefore, external knowledge sources can be used to provide additional information. We adopt ConceptNet \citep{conceptnet} as our knowledge graph $G=(V, E)$, which contains over 8 million entities as nodes and over 21 million relations as edges. In the following, we use triple to refer to two neighboring nodes and the edge connecting them, i.e. $(u\in V, p=(u, v)\in E, v\in V)$, with $u$ being the subject, $p$ the relation, and $v$ the object.

Suppose the question mentions an entity $e_q\in V$ and the choice contains an entity $e_c\in V$\footnote{CommonsenseQA provides the question/choice entity. For OpenBookQA, we choose from the extracted entities that are most frequent in retrieved facts. See Appendix for details.}.
We then employ the KCR method \citep{KCR} to select relation triples. If there is a direct edge $r$ from $e_q$ to $e_c$ in $G$, we choose this triple ($e_q$, $r$, $e_c$). Otherwise, we retrieve all the $N$ triples containing $e_c$. Each triple $j$ is assigned a score $s_j$ which is the product of its triple weight $w_j$ provided by ConceptNet and relation type weight $t_{r_j}$:

\begin{align}
    s_j &= w_j \cdot t_{r_j} = w_j\cdot \frac{N}{N_{r_j}}
\end{align}

Here, $r_j$ is the relation type of the triple $j$, and $N_{r_j}$ is the number of triples among the retrieved triples that have the relation type $r_j$. In other words, this process favors rarer relation types. Finally, the triple with the highest weight is chosen. 

\begin{figure*}[t]
    \centering
\includegraphics[width=0.99\textwidth,trim=3.6cm 1.8cm 3.0cm 4.2cm]{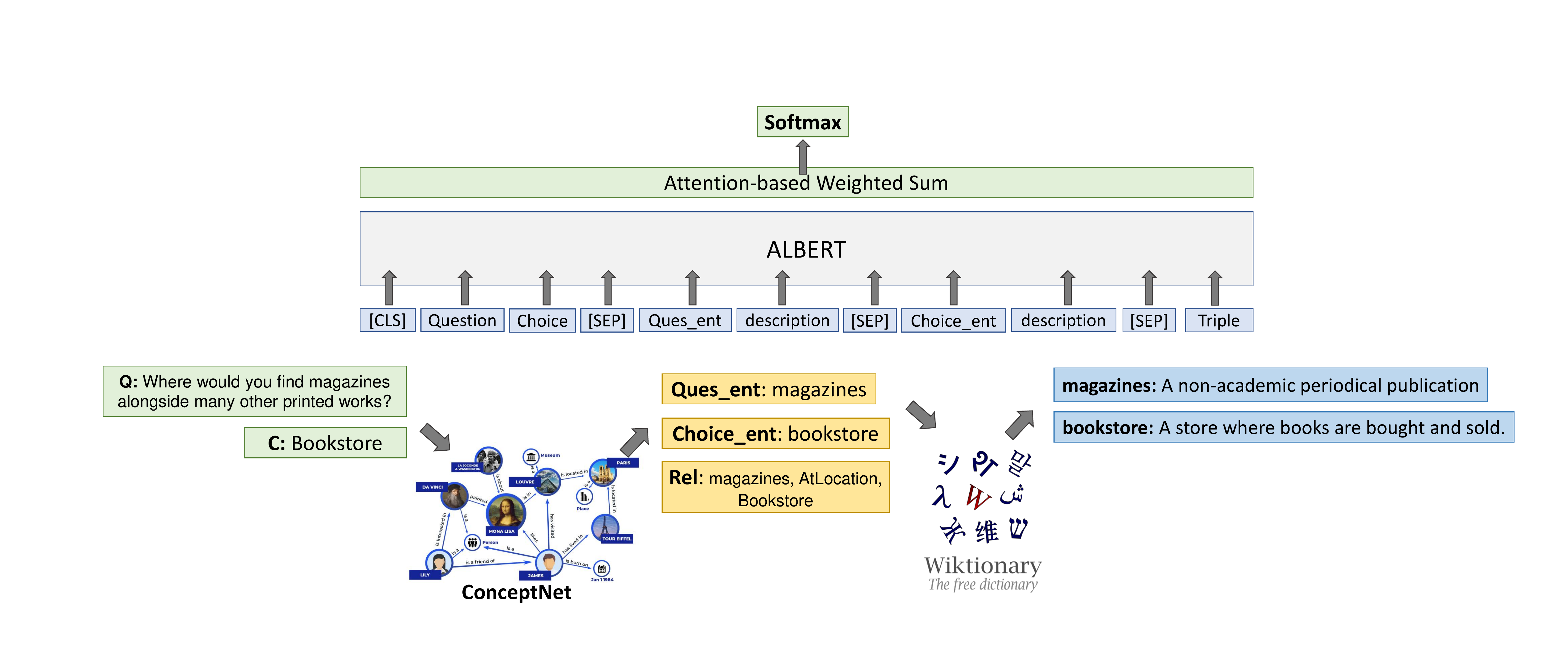}
\caption{In our model, the input to ALBERT includes the question, choice, entity names, description text and triple. An attention-based weighted sum and a softmax layer process the output from ALBERT to produce the prediction.}
\label{fig:archi}
\end{figure*}

\subsection{Contextual information}
\label{sec:context}
The retrieved entities and relations from the knowledge graph are described by their surface form. Without additional context, it is hard for the language model to understand its exact meaning, especially for proper nouns.

Therefore, we leverage large-scale online dictionaries to provide definitions as context. We use a dump of Wiktionary\footnote{\url{https://www.wiktionary.org/}} which includes definitions of 999,614 concepts. For every concept, we choose its first definition entry in Wiktionary as the description. For every question/choice concept, we find its closest match in Wiktionary by using the following forms in order: i) original form; ii) lemma form by Spacy \citep{spacy}; iii) base word (last word).
For example, the concept ``taking notes" does not appear in its original form in Wiktionary, but its lemma form ``take notes'' is in Wiktionary and we get its description text: ``\textit{To make a record of what one hears or observes for future reference}''. In this way, we find descriptions of all entities in our experiments. The descriptions of the question and choice concept are denoted by $d_q$ and $d_c$, respectively.

Finally, we feed the question, choice, descriptions and triple (from Section \ref{sec:form}) into the ALBERT model \citep{ALBERT} in the following format:
[CLS] $q$ $c$ [SEP] $e_q$: $d_q$ [SEP] $e_c$: $d_c$ [SEP] triple.


\vspace{-2mm}
\subsection{Reasoning}
\label{sec:reason}
On top of ALBERT, we leverage an attention-based weighted sum and a softmax layer to generate the relevance score for the question-choice pair. In detail, suppose the output representations of ALBERT is $(\bx_0, ..., \bx_m)$, where $\bx_i\in R^d$. We compute a weighted sum of these embeddings based on attention:
\begin{align}
    q_i &= \bu^T\bx_i \\
    \alpha_i &= \mbox{softmax}(q_i) \\
    \bv &= \sum_{i=0}^m{\alpha_i \bx_i},
\end{align}
where $\bu$ is a parameter vector. The relevance score between the question and the choice is then $s=\mbox{softmax}(\bv^T\bb)$, where $\bb\in R^d$ is a parameter vector and the softmax is computed over all choices for the cross-entropy loss function.

The architecture of our model DEKCOR and the construction of input is shown in Fig.~\ref{fig:archi}.

\begin{table}[tb]
\centering
\caption{Statistics of CommonsenseQA (CSQA) and OpenBookQA (OBQA).}
\label{tab:statistics}
\begin{tabular}{lcccc}
\thickhline
\textbf{Dataset}& \textbf{Train} & \textbf{Dev}  & \textbf{Test} & \textbf{Choices}         \\
\hline
CSQA& 9,741 & 1,221 & 1,140 & 5\\
OBQA& 4,957 & 500 & 500 & 4\\
\thickhline
\end{tabular}
\end{table}

\begin{table}[tb]
\centering
\caption{Accuracy on the test set of CommonsenseQA.}
\label{tab:csqa}
\begin{tabular}{lcc}
\thickhline
\textbf{Methods}& \textbf{Single} & \textbf{Ensemble}           \\
\hline
BERT+OMCS & 62.5 & - \\
RoBERTa & 72.1 & 72.5 \\
RoBERTa+HyKAS&73.2&-\\
XLNet+DREAM&-&73.3\\
RoBERTa+KE & 73.3 & -\\
RoBERTa+KEDGN &-& 74.4\\
XLNet+GraphReason & 75.3&-\\
ALBERT & -&76.5 \\ 
RoBERTa+MHGRN & 75.4 & 76.5\\ 
ALBERT+PG-Full & 75.6 & 78.2\\ 
T5 & 78.1 & - \\ 
ALBERT+KRD & 78.4 & - \\
UnifiedQA & 79.1 & -\\
ALBERT+KCR &79.5 &-\\
\hline
DEKCOR (ours) & \textbf{80.7} & \textbf{83.3}\\
\thickhline
\end{tabular}
\end{table}

\begin{table}[tb]
\centering
\caption{Accuracy on the test set of OpenBookQA.}
\label{tab:obqa}
\begin{tabular}{lc}
\thickhline
\textbf{Methods}& \textbf{Accuracy}          \\
\hline
BERT + Careful Selection & 72.0 \\
AristoRoBERTa & 77.8 \\ 
ALBERT + KB & 81.0 \\
ALBERT + PG-Full & 81.8\\
TTTTT (T5-3B) & 83.2\\
UnifiedQA (T5-11B) & \textbf{87.2}\\
\hline
DEKCOR (ours) & \underline{82.4}\\
\thickhline
\end{tabular}
\end{table}

\begin{table}[t]
\centering
\caption{Ablation results on the dev sets of CommonsenseQA and OpenBookQA.}
\label{tab:ablation}
\begin{tabular}{lccc}
\thickhline
\textbf{Methods}& \textbf{CSQA} & \textbf{OBQA}\\
\hline
DEKCOR & 84.7 & 82.2\\
Triple Only & 82.0 & 80.0 \\
Description Only & 80.3 & 81.8\\
No Context & 78.9 & 80.0\\
\thickhline
\end{tabular}
\end{table}

\section{Experiments}
\label{sec:exp}
\subsection{Datasets and baselines}
We evaluate our model on two benchmark datasets of multiple-choice questions for commonsense question answering: CommonsenseQA \citep{csqa} and OpenBookQA \citep{mihaylov2018can}. CommonsenseQA creates questions from ConceptNet entities and relations; OpenBookQA probes elementary science knowledge from a book of 1,326 facts. 
The statistics of the datasets is provided in Table \ref{tab:statistics}. For OpenBookQA, we follow prior approaches \citep{wang2020connecting} to append top 5 retrieved facts provided by Aristo team \citep{clark2019f} to the input. We also pre-train our OpenBookQA model on CommonsenseQA's training set as we find it helps to boost the performance.

We compare our models with state-of-the-art baselines, which all employ pre-trained models including RoBERTa \cite{roberta}, XLNet \cite{xlnet}, ALBERT \cite{ALBERT} and T5 \cite{t5} and some adopt additional modules to process knowledge information. A detailed description of the baselines is in the Appendix.

\subsection{Results}
\textbf{CommonsenseQA.} Table~\ref{tab:csqa} shows the accuracy on the test set of CommonsenseQA. For a fair comparison, we categorize the results into single models and ensemble models. Our ensemble model consists of 7 single models with different initialization random seeds, and its output is the majority of choices selected by these single models. More implementation details are shown in the Appendix.

Our proposed DEKCOR outperforms the previous state-of-the-art result by 1.2\% (single model) and 3.8\% (ensemble model). This demonstrates the effectiveness of the usage of knowledge description to provide context.

Furthermore, we notice two trends based on the results. First, the underlying pre-trained language model is important in commonsense QA quality. In general, we observe this order of accuracy among these language models: BERT$<$RoBERTa$<$XLNet$<$ALBERT$<$T5. Second, the additional knowledge module is critical to provide external information for reasoning. For example, RoBERTa+KEDGN outperforms the vanilla RoBERTa by 1.9\%, and our model outperforms the vanilla ALBERT model by 6.8\% in accuracy.

\noindent\textbf{OpenBookQA.} Table~\ref{tab:obqa} shows the test set accuracy on OpenBookQA. All results are from single models. Note that the two best-performing models, i.e. UnifiedQA \citep{unifiedqa} and TTTTT \citep{t5}, are based on the T5 generation model, with 11B and 3B parameters respectively. Thus, they are computationally very expensive. Except these T5-based systems, DEKCOR achieves the best accuracy among all baselines.

\noindent\textbf{Ablation study}. Table~\ref{tab:ablation} shows that the usage of concept descriptions from Wiktionary and triple from ConceptNet can help improve the accuracy of DEKCOR on the dev set of CommonsenseQA by 2.7\% and 4.4\% respectively. We observe similar results on OpenBookQA. This demonstrates that additional context information can help with fusing knowledge graph into language modeling for commonsense question answering.
\begin{table}[t]
\renewcommand{\arraystretch}{1.1}
\begin{small}
\begin{tabular}{p{7.5cm}}
\hline
\textbf{CommonsenseQA Question:} \\
Bats have many quirks, with the exception of \rule{0.5cm}{0.15mm} . \\
\textbf{Question entity description:} \\
bat: Any of the flying mammals of the order Chiroptera, usually small and nocturnal, insectivorous or frugivorous.
\\ \hline
\textbf{Model w/o description chooses:} eating bugs\\
    \textbf{Model w/ description chooses:} laying eggs\\\hhline{=}
\textbf{OBQA Question:}\\                                                 
Alligators  \rule{0.5cm}{0.15mm} . \\
\textbf{Question entity description:} \\
alligator: Either of two species of large amphibious reptile, ..., which have sharp teeth and very strong jaws...
\\ \hline
\textbf{Model w/o description chooses:} eat gar\\
    \textbf{Model w/ description chooses:} are warm-blooded\\
\hline
\end{tabular}
\end{small}
\caption{Examples from CommonsenseQA and OBQA dataset showing the effectiveness of entity descriptions. }\label{tab:examples}
\end{table}

\noindent\textbf{Case Study.} Table \ref{tab:examples} shows two examples from CommonsenseQA and OBQA respectively. In the first example, without additional description the model would not know relevant information about bats, like they are insectivorous, leading to the wrong answer ``eating bugs''. With the description, the model knows that bats eat bugs, so it chooses ``laying eggs'' as the answer. Similarly, for the second question, the ``sharp teeth and very strong jaws'' in the description hint that alligators are likely carnivorous, and reptiles are likely cold-blooded. The entity description leads to the correct answer of ``eat gar''.


\section{Conclusions}
\label{sec:conclusion}
In this paper, we propose to fuse context information into knowledge graphs for commonsense question answering. As a knowledge graph often lacks descriptions for the contained entities and relations, we leverage Wiktionary to provide definitive text for each entity as additional input to the pre-trained language model ALBERT. The resulting DEKCOR model achieves state-of-the-art results on the benchmark datasets CommonsenseQA and OpenBookQA. Ablation studies demonstrate the effectiveness of the proposed usage of knowledge description and knowledge triple information in commonsense question answering.

\section*{Acknowledgements}
We thank the anonymous reviewers for their valuable comments.

\bibliography{acl2021}
\bibliographystyle{acl_natbib}

\clearpage
\appendix
\section{Implementation Details}

\noindent \textbf{Identification of $e_q$ and $e_c$.} CommonsenseQA specifies the question entity in each question and each answer choice is also an entity in ConceptNet. We use them as $e_q$ and $e_c$. For OpenBookQA, we identify all ConceptNet entities in the question and answer text and count their number of occurrences in the retrieved text. For a triple $(e_q, r, e_c)$, we define its weight as $n_{e_q}+n_{e_c}$, where $n_e$ is the number of occurrences in retrieved text. The edge with the largest weight is picked. If no edge is found between question and answer entities, we use the answer entity with the most occurrences to find triples. For Wiktionary descriptions, we find descriptions for $e_q$ and $e_c$ with the most occurrences as well.

\noindent \textbf{Using ConceptNet.} Since ConceptNet contains a lot of weak relations, we only use the following relations for our triples: CausesDesire, HasProperty, CapableOf, PartOf, AtLocation, Desires, HasPrerequisite, HasSubevent, Antonym, Causes.

\noindent\textbf{Optimization.}
We use the AdamW \citep{adamw} optimizer with a learning rate of 2e-5. The batch size is 8. We limit the maximum length of the input sequence to 192 tokens. The model is trained for 10 epochs. We use the Huggingface \citep{huggingface} implementation for the ALBERT model.

\section{Baseline Methods}
 GraphReason \citep{lv2020graph} retrieves knowledge from both structured knowledge base and plain text. \\ 
 PG-FULL \citep{wang2020connecting} fine-tunes GPT-2 on ConceptNet to generate knowledgeable paths between knowledge graph concepts. \\
UnifiedQA \citep{unifiedqa} pre-trains T5 on a variety of QA datasets for general QA tasks.\\
MHGRN \citep{feng2020scalable} adopts the multi-hop graph relation network to perform reasoning.\\ 
HyKAS \citep{ma2019towards} employs an option comparison network to consume ConceptNet triples.\\
ALBERT+KRD retrieves commonsense knowledge from Open Mind Common Sense and then uses a self-attention module to compute a weighted sum of these triple representations. \\
BERT + Selection \cite{Banerjee2019CarefulSO} improves the result on OpenBookQA via abductive information retrieval , information gain based re-ranking, passage selection and weighted scoring. \\ 
ALBERT+KB also improves retrieval results on OpenBookQA by token-based and embedding-based retrieval.
TTTTT \citep{t5} fine-tunes the T5 language generation model on OpenBookQA.
\end{document}